\documentclass{article}

\usepackage[english]{babel}

\usepackage[letterpaper,top=2cm,bottom=2cm,left=3cm,right=3cm,marginparwidth=1.75cm]{geometry}

\usepackage{amsmath}
\usepackage{graphicx}
\usepackage[colorlinks=true, allcolors=blue]{hyperref}
\usepackage{adjustbox}
\usepackage{xcolor}
\usepackage{multirow}
\usepackage{subcaption}
\usepackage{array}
\usepackage{makecell}
\usepackage{bbding}
\usepackage{algorithm}
\usepackage{algorithmic}
\usepackage{amsmath,amssymb}

\definecolor{light-gray}{gray}{0.95}
\newcommand{\code}[1]{\colorbox{light-gray}{\texttt{#1}}}
\title{A Benchmark of Long-tailed Instance Segmentation with Noisy Labels}
\author{Guanlin Li \\ S-Lab, NTU \\ guanlin001@e.ntu.edu.sg
 \and Guowen Xu \\ City University of Hong Kong \\ guowenxu@cityu.edu.hk \and Tianwei  Zhang \\ NTU \\ tianwei.zhang@ntu.edu.sg}

\begin{document}
\maketitle

\begin{abstract}

In this paper, we consider the instance segmentation task on a long-tailed dataset, which contains label noise, i.e., some of the annotations are incorrect. There are two main reasons making this case realistic. First, datasets collected from real world usually obey a long-tailed distribution. Second, for instance segmentation datasets, as there are many instances in one image and some of them are tiny, it is easier to introduce noise into the annotations. Specifically, we propose a new dataset, which is a large vocabulary long-tailed dataset containing label noise for instance segmentation. Furthermore, we evaluate previous proposed instance segmentation algorithms on this dataset. The results indicate that the noise in the training dataset will hamper the model in learning rare categories and decrease the overall performance, and inspire us to explore more effective approaches to address this practical challenge. The code and dataset are available in \url{https://github.com/GuanlinLee/Noisy-LVIS}.
\end{abstract}

\section{Introduction}

Instance segmentation models~\cite{he_mask_2017,wang2020solov2} provide each pixel in an image a label to indicate the specific semantic information, which can be seen as a finer inference of object detection models~\cite{ren2015faster,liu2016ssd} and semantic segmentation models~\cite{badrinarayanan2017segnet,chen2018encoder}. Most of previous works focus on exploring more effective algorithms on balanced and clean datasets, such as MS COCO~\cite{mscoco}. However, instances included in images usually obey a long-tailed distribution~\cite{lin_focal_2017, cao_learning_2019,cui_class-balanced_2019}, which means 
most of the instances belong to a small part of categories, which are called ``head'' classes, and few of instances belong to other categories, which are called ``body'' and ``tail'' classes. If users do not modify the training algorithms to fit an unbalanced dataset, the model will give higher confidence to the instances from head classes~\cite{lin_focal_2017}, which harms the generalizability to the instances from body and tail classes, causing accuracy decrease. On the other hand, the annotations for each item in the dataset are usually given by people who are paid by the big companies, like Google\footnote{https://venturebeat.com/ai/probeat-google-still-needs-you-to-label-photos-for-its-ml/} and Meta\footnote{https://www.reuters.com/article/us-facebook-ai-focus-idUSKCN1SC01T}. In reality, these annotators are not always experts~\cite{whitehill_whose_2009}, as the companies wish to decrease the cost. So, there are usually some wrong labels in the dataset, called label noise. For example, in Clothing1M~\cite{xiao_learning_2015} the percentage of label noise is about 38.5\%~\cite{song_learning_2022}, and in WebVision~\cite{webvision} the label noise ratio is about 20.0\%~\cite{song_learning_2022}. If users do not adopt any approach to mitigate the impact of the label noise, the accuracy of the model will significantly decrease on the clean test set, because the incorrect labels can provide wrong supervised information during the training process. So, there exists a gap which can cause performance drop when users train a model on a personal dataset, which is unbalanced or contains label noise. To address the challenges in training a model on long-tailed datasets or noisy datasets, many studies have been published, recently.

Similar to exploring how to train classifiers on unbalanced datasets, solving long-tailed challenges in the instance segmentation task raises the attention of the deep learning community. Different from the classification task on long-tailed datasets, backgrounds in the instance segmentation datasets always occupy an overwhelming position, and in foreground, there are head classes and tail classes, which means blindly adopting approaches in classification tasks cannot address the problem in instance segmentation. So, there are a series of new methods~\cite{hu_learning_2020,hsieh_droploss_2021,wang_Seesaw_2021} proposed in solving long-tailed instance segmentation task, including new data augmentation methods, new model architectures and new loss functions. These solutions can work well on long-tailed datasets, without considering the label noise. However, as aforementioned challenges in deep learning, noise labels are more popular in instance segmentation datasets, because there are more instances in the datasets and annotating them exhaustively and precisely is more difficult. Specifically, the noise in instance segmentation datasets source from two aspects, i.e., foreground instances missed annotations (false negative background) and foreground instances mis-annotated into other classes (noisy label). In this paper, we focus on the latter one, which is a more harmful type of noise in the dataset. There are numerous studies focusing on addressing the noisy label problem in classification tasks, such as new model structures~\cite{sukhbaatar_learning_2015} and new loss functions~\cite{mae}. And Yang at el.~\cite{lncis} propose
a two-stage training fashion to overcome the label noise in the instance segmentation task. On the balanced dataset containing label noise, adopting such methods can successfully overcome the accuracy drop. However, they do not consider a more practical setting, i.e., a long-tailed noisy dataset.

Recently, to fill the gap and make a connection between long-tailed recognition and noisy label learning, several works~\cite{twrl,sklf} combine long-tailed learning methods and technologies for solving label noise. The results indicate that this direct solution is helpful to train a model on an unbalanced dataset containing label noise. However, in the instance segmentation task, background objects and foreground objects are required to be handled separately, which is different from the image recognition task, causing that previous methods for long-tailed recognition and noisy label learning cannot be applied to this task directly. On the other hand, we find a lack of relevant datasets, which are both long-tailed and noisy, in the instance segmentation task. It brings the difficulties for other researchers to explore approaches to address these practical challenges. So, in this paper, we attempt to introduce a benchmark for the instance segmentation task on long-tailed and noisy datasets.

Formally, there are two main contributions in this paper, i.e., introducing a new long-tailed noisy dataset and systematically evaluating previously proposed methods on this dataset. Specifically, we build a noisy long-tailed datasets from the LVIS v1~\cite{gupta_lvis_2019}, which is a large vocabulary long-tailed instance segmentation dataset containing 1,203 different categories for the foreground objects. We will introduce the details of how we add noisy labels into the LVIS v1 in Section~\ref{sec:nlvis}. For our second contribution, we evaluate EQL~\cite{tan_equalization_2020}, DropLoss~\cite{hsieh_droploss_2021} and Seesaw Loss~\cite{wang_Seesaw_2021} on this new dataset under different noise settings and samplers in Section~\ref{sec:exp}. More evaluation results will be added in the future. From the results under different noise settings and samplers, we find that previous proposed long-tailed instance segmentation losses cannot keep robust under label noise and data sampling methods have a significant impact on the final results, due to distribution shift caused by the label noise. We hope this benchmark and the new proposed dataset can shrink the gap between ideal experimental settings and practical datasets and bring guides for researchers in further exploration.

\section{Related Works}

\subsection{LVIS v1}

LVIS v1 is a dataset for Large Vocabulary Instance Segmentation, derived from the Microsoft Common Objects in Context (MS COCO)~\cite{mscoco} with much more fine-grained annotations, which means that the images in LVIS v1 are the same as them in the MS COCO, but the categories of each instance in the images are more precise. Specifically, there are 1,203 different classes in LVIS v1, and in MS COCO, there are only 80 classes. As a large vocabulary dataset, the evaluation process must be accurate. To achieve it, the LVIS v1 adopts positive sets and negative sets for each category\footnote{This can cause that some instances are not annotated and are treated as background. In this paper, we do not consider the false negative samples' influence in the training procedure.}. As for the categories in LVIS v1, each of them is a WordNet~\cite{wordnet} \textit{synset}, i.e., a word sense specified by a set of synonyms and a definition. Ideally, we can treat the LVIS v1 as an exhaustive and clean long-tailed dataset. And, we will explore the previous long-tailed segmentation methods trained on a noisy dataset, i.e., adding symmetry and asymmetry noise labels into the training set of the LVIS v1.

\subsection{Long-tailed Instance Segmentation}

Compared with the long-tailed object classification task~\cite{ren_balanced_2020,menon_long-tail_2021}, the long-tailed instance segmentation task is much closer to the long-tailed object detection task~\cite{tan_equalization_2020}. The reason is that the object detection task can be seen as a sub-task in the instance segmentation task, as some instance segmentation pipeline first generates bounding boxes for each object and then generates masks inside the bounding boxes. Because instance segmentation models need to recognize backgrounds and foregrounds, and generate masks for each foreground object, background-foreground classification plays a key role. However, backgrounds occupy an overwhelming number in the dataset, which means that simply adopting solutions proposed in long-tailed object classification tasks cannot address the background-foreground classification challenges. In long-tailed instance segmentation tasks, there are several new proposed approaches, such as gradient calibration~\cite{tan_equalization_2020, wang_Seesaw_2021}, two-stage training~\cite{wang_devil_2020}, data augmentation~\cite{ghiasi_simple_2021}, and new model structures~\cite{zang_fasa_2021}. In this paper, we mainly consider evaluating loss functions designed for long-tailed instance segmentation tasks.

\subsection{Noisy Label Learning}

Various approaches have been proposed to address the label noise from different perspectives, including new model architectures~\cite{sukhbaatar_learning_2015}, regularization terms~\cite{bilevel}, robust loss functions~\cite{mae}, loss adjustment~\cite{bootstrapping} and sample selection~\cite{coteaching}. \cite{sukhbaatar_learning_2015} proposed a new model architecture containing a noise adaptation layer to model the label transition pattern with a noise transition matrix. Furthermore, a generative classification layer is found to be more robust against the label noise~\cite{rog}. On the other hand, both adversarial training~\cite{goodfellow_explaining_2015} and label smoothing~\cite{labelsmooth}, which could be seen as regularization, were found to be efficient at improving the model's tolerance to label noise. For the robust loss functions, \cite{mae} proved that the mean absolute error (MAE) loss was robust to the label noise, but it harmed the model's generalizability. So, the generalized cross entropy (GCE)~\cite{zhang_generalized_2018} keeping the advantages from cross-entropy loss and MAE loss is proposed. \cite{patrini_making_2017} proposed a Backward correction method and a Forward correction method to address the label noise challenge by adjusting the losses.  Furthermore, Bootstrapping~\cite{bootstrapping} was the first label refurbishment method to automatically correct the data labels during the training process. For the sample selection, \cite{coteaching} adopted two models to adaptively choose samples to train its peer model based on the value of the training loss, which was called the Co-teaching method. Combining with semi-supervised learning, DivideMix~\cite{li_dividemix_2020} trained models on clean dataset and noisy dataset.

Although there are numerous researches focusing on solving the label noise problem, most of them are designed for balanced datasets and the image classification task. In this paper, we first systematically evaluate previous methods of the instance segmentation task on a noisy long-tailed dataset.

\section{LVIS v1 with Noise}
\label{sec:nlvis}

In this section, we mainly analyze how to add asymmetry (class-related) noise into the training set of the LVIS v1. Because there are 1203 categories, analyzing the semantics of each category manually and gathering classes with similar semantics manually can be impracticable and require a lot of workloads. We create an automated toolkit, to analyze the semantics based on WordNet. We will introduce the analyzing process in the follows.

\subsection{Parse synset Categories}

\begin{figure}
    \centering
    \adjincludegraphics[height=500pt, trim={{.9\width} 0 0 0}, clip, angle=90]{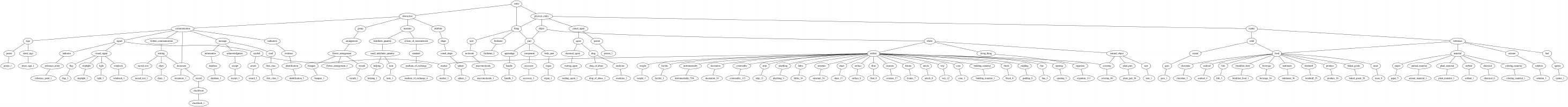}
    \caption{}
\label{fig:tree}
\end{figure}

To parse the synset categories in the LVIS v1, we adopt \code{nltk}~\cite{bird2009natural}, which is a natural language toolkit. With it, we build a tree to represent the relations between all categories in the LVIS v1, which can be found in Figure~\ref{fig:tree}. Specifically, the numbers in the child nodes represent how many synset categories belong to this node, i.e., their hypernyms are this node\footnote{We do not plot the categories as the child nodes, because it needs a tree with depth 18. But the toolkit we provide can plot a tree with arbitrary depths.}. We manually pick 25 hypernyms as super classes and one ``others'' as a super class to contain the remaining categories, which do not belong to any of the 25 super classes. Then, we divide categories based on their hypernyms into each super class, and there is no overlap between different super classes. The details of the super classes and the number of categories in each super class can be found in Table~\ref{tab:sc}.

\begin{table}[]
\centering
\begin{minipage}{\linewidth}
\begin{adjustbox}{max width=1.0\linewidth}
\begin{tabular}{c|cccc|cccc|cccc|cccc}
\hline
\textbf{Name of Super Class} & \multicolumn{4}{c|}{instrumentality} & \multicolumn{4}{c|}{food} & \multicolumn{4}{c|}{living\_thing} & \multicolumn{4}{c}{commodity} \\ \hline
\multirow{2}{*}{\textbf{Number of Categories}} & \multicolumn{1}{c|}{\textbf{r}} & \multicolumn{1}{c|}{\textbf{c}} & \multicolumn{1}{c|}{\textbf{f}} & \textbf{total} & \multicolumn{1}{c|}{\textbf{r}} & \multicolumn{1}{c|}{\textbf{c}} & \multicolumn{1}{c|}{\textbf{f}} & \textbf{total} & \multicolumn{1}{c|}{\textbf{r}} & \multicolumn{1}{c|}{\textbf{c}} & \multicolumn{1}{c|}{\textbf{f}} & \textbf{total} & \multicolumn{1}{c|}{\textbf{r}} & \multicolumn{1}{c|}{\textbf{c}} & \multicolumn{1}{c|}{\textbf{f}} & \textbf{total} \\ \cline{2-17} 
 & \multicolumn{1}{c|}{150} & \multicolumn{1}{c|}{200} & \multicolumn{1}{c|}{174} & 524 & \multicolumn{1}{c|}{52} & \multicolumn{1}{c|}{58} & \multicolumn{1}{c|}{29} & 139 & \multicolumn{1}{c|}{34} & \multicolumn{1}{c|}{60} & \multicolumn{1}{c|}{19} & 113 & \multicolumn{1}{c|}{23} & \multicolumn{1}{c|}{40} & \multicolumn{1}{c|}{50} & 113 \\ \hline
\end{tabular}
\end{adjustbox}\\
\\

\begin{adjustbox}{max width=1.0\linewidth}
\begin{tabular}{c|cccc|cccc|cccc|cccc}
\hline
\textbf{Name of Super Class} & \multicolumn{4}{c|}{natural\_object} & \multicolumn{4}{c|}{decoration} & \multicolumn{4}{c|}{structure} & \multicolumn{4}{c}{communication} \\ \hline
\multirow{2}{*}{\textbf{Number of Categories}} & \multicolumn{1}{c|}{\textbf{r}} & \multicolumn{1}{c|}{\textbf{c}} & \multicolumn{1}{c|}{\textbf{f}} & \textbf{total} & \multicolumn{1}{c|}{\textbf{r}} & \multicolumn{1}{c|}{\textbf{c}} & \multicolumn{1}{c|}{\textbf{f}} & \textbf{total} & \multicolumn{1}{c|}{\textbf{r}} & \multicolumn{1}{c|}{\textbf{c}} & \multicolumn{1}{c|}{\textbf{f}} & \textbf{total} & \multicolumn{1}{c|}{\textbf{r}} & \multicolumn{1}{c|}{\textbf{c}} & \multicolumn{1}{c|}{\textbf{f}} & \textbf{total} \\ \cline{2-17} 
 & \multicolumn{1}{c|}{9} & \multicolumn{1}{c|}{15} & \multicolumn{1}{c|}{14} & 38 & \multicolumn{1}{c|}{5} & \multicolumn{1}{c|}{10} & \multicolumn{1}{c|}{9} & 24 & \multicolumn{1}{c|}{8} & \multicolumn{1}{c|}{7} & \multicolumn{1}{c|}{9} & 24 & \multicolumn{1}{c|}{7} & \multicolumn{1}{c|}{9} & \multicolumn{1}{c|}{7} & 23 \\ \hline
\end{tabular}
\end{adjustbox}\\
\\

\begin{adjustbox}{max width=1.0\linewidth}
\begin{tabular}{c|cccc|cccc|cccc|cccc}
\hline
\textbf{Name of Super Class} & \multicolumn{4}{c|}{creation} & \multicolumn{4}{c|}{material} & \multicolumn{4}{c|}{fabric} & \multicolumn{4}{c}{sheet} \\ \hline
\multirow{2}{*}{\textbf{Number of Categories}} & \multicolumn{1}{c|}{\textbf{r}} & \multicolumn{1}{c|}{\textbf{c}} & \multicolumn{1}{c|}{\textbf{f}} & \textbf{total} & \multicolumn{1}{c|}{\textbf{r}} & \multicolumn{1}{c|}{\textbf{c}} & \multicolumn{1}{c|}{\textbf{f}} & \textbf{total} & \multicolumn{1}{c|}{\textbf{r}} & \multicolumn{1}{c|}{\textbf{c}} & \multicolumn{1}{c|}{\textbf{f}} & \textbf{total} & \multicolumn{1}{c|}{\textbf{r}} & \multicolumn{1}{c|}{\textbf{c}} & \multicolumn{1}{c|}{\textbf{f}} & \textbf{total} \\ \cline{2-17} 
 & \multicolumn{1}{c|}{6} & \multicolumn{1}{c|}{4} & \multicolumn{1}{c|}{7} & 17 & \multicolumn{1}{c|}{5} & \multicolumn{1}{c|}{5} & \multicolumn{1}{c|}{5} & 15 & \multicolumn{1}{c|}{1} & \multicolumn{1}{c|}{6} & \multicolumn{1}{c|}{7} & 14 & \multicolumn{1}{c|}{1} & \multicolumn{1}{c|}{3} & \multicolumn{1}{c|}{9} & 13 \\ \hline
\end{tabular}
\end{adjustbox}\\
\\

\begin{adjustbox}{max width=1.0\linewidth}
\begin{tabular}{c|cccc|cccc|cccc|cccc}
\hline
\textbf{Name of Super Class} & \multicolumn{4}{c|}{way} & \multicolumn{4}{c|}{strip} & \multicolumn{4}{c|}{plaything} & \multicolumn{4}{c}{padding} \\ \hline
\multirow{2}{*}{\textbf{Number of Categories}} & \multicolumn{1}{c|}{\textbf{r}} & \multicolumn{1}{c|}{\textbf{c}} & \multicolumn{1}{c|}{\textbf{f}} & \textbf{total} & \multicolumn{1}{c|}{\textbf{r}} & \multicolumn{1}{c|}{\textbf{c}} & \multicolumn{1}{c|}{\textbf{f}} & \textbf{total} & \multicolumn{1}{c|}{\textbf{r}} & \multicolumn{1}{c|}{\textbf{c}} & \multicolumn{1}{c|}{\textbf{f}} & \textbf{total} & \multicolumn{1}{c|}{\textbf{r}} & \multicolumn{1}{c|}{\textbf{c}} & \multicolumn{1}{c|}{\textbf{f}} & \textbf{total} \\ \cline{2-17} 
 & \multicolumn{1}{c|}{4} & \multicolumn{1}{c|}{3} & \multicolumn{1}{c|}{5} & 12 & \multicolumn{1}{c|}{0} & \multicolumn{1}{c|}{5} & \multicolumn{1}{c|}{6} & 11 & \multicolumn{1}{c|}{4} & \multicolumn{1}{c|}{1} & \multicolumn{1}{c|}{4} & 9 & \multicolumn{1}{c|}{1} & \multicolumn{1}{c|}{2} & \multicolumn{1}{c|}{5} & 8 \\ \hline
\end{tabular}
\end{adjustbox}\\
\\

\begin{adjustbox}{max width=1.0\linewidth}
\begin{tabular}{c|cccc|cccc|cccc|cccc}
\hline
\textbf{Name of Super Class} & \multicolumn{4}{c|}{fixture} & \multicolumn{4}{c|}{surface} & \multicolumn{4}{c|}{measure} & \multicolumn{4}{c}{causal\_agent} \\ \hline
\multirow{2}{*}{\textbf{Number of Categories}} & \multicolumn{1}{c|}{\textbf{r}} & \multicolumn{1}{c|}{\textbf{c}} & \multicolumn{1}{c|}{\textbf{f}} & \textbf{total} & \multicolumn{1}{c|}{\textbf{r}} & \multicolumn{1}{c|}{\textbf{c}} & \multicolumn{1}{c|}{\textbf{f}} & \textbf{total} & \multicolumn{1}{c|}{\textbf{r}} & \multicolumn{1}{c|}{\textbf{c}} & \multicolumn{1}{c|}{\textbf{f}} & \textbf{total} & \multicolumn{1}{c|}{\textbf{r}} & \multicolumn{1}{c|}{\textbf{c}} & \multicolumn{1}{c|}{\textbf{f}} & \textbf{total} \\ \cline{2-17} 
 & \multicolumn{1}{c|}{1} & \multicolumn{1}{c|}{0} & \multicolumn{1}{c|}{6} & 7 & \multicolumn{1}{c|}{4} & \multicolumn{1}{c|}{2} & \multicolumn{1}{c|}{0} & 6 & \multicolumn{1}{c|}{3} & \multicolumn{1}{c|}{2} & \multicolumn{1}{c|}{1} & 6 & \multicolumn{1}{c|}{1} & \multicolumn{1}{c|}{2} & \multicolumn{1}{c|}{2} & 5 \\ \hline
\end{tabular}
\end{adjustbox}\\
\\

\begin{adjustbox}{max width=1.0\linewidth}
\begin{tabular}{c|cccc|cccc|cccc|cccc}
\hline
\textbf{Name of Super Class} & \multicolumn{4}{c|}{block} & \multicolumn{4}{c|}{float} & \multicolumn{4}{c|}{line} & \multicolumn{4}{c}{opening} \\ \hline
\multirow{2}{*}{\textbf{Number of Categories}} & \multicolumn{1}{c|}{\textbf{r}} & \multicolumn{1}{c|}{\textbf{c}} & \multicolumn{1}{c|}{\textbf{f}} & \textbf{total} & \multicolumn{1}{c|}{\textbf{r}} & \multicolumn{1}{c|}{\textbf{c}} & \multicolumn{1}{c|}{\textbf{f}} & \textbf{total} & \multicolumn{1}{c|}{\textbf{r}} & \multicolumn{1}{c|}{\textbf{c}} & \multicolumn{1}{c|}{\textbf{f}} & \textbf{total} & \multicolumn{1}{c|}{\textbf{r}} & \multicolumn{1}{c|}{\textbf{c}} & \multicolumn{1}{c|}{\textbf{f}} & \textbf{total} \\ \cline{2-17} 
 & \multicolumn{1}{c|}{2} & \multicolumn{1}{c|}{1} & \multicolumn{1}{c|}{1} & 4 & \multicolumn{1}{c|}{1} & \multicolumn{1}{c|}{1} & \multicolumn{1}{c|}{2} & 4 & \multicolumn{1}{c|}{0} & \multicolumn{1}{c|}{2} & \multicolumn{1}{c|}{1} & 3 & \multicolumn{1}{c|}{0} & \multicolumn{1}{c|}{1} & \multicolumn{1}{c|}{2} & 3 \\ \hline
\end{tabular}
\end{adjustbox}\\
\\

\begin{adjustbox}{max width=1.0\linewidth}
\begin{tabular}{c|cccc|cccc|cccc}
\hline
\textbf{Name of Super Class} & \multicolumn{4}{c|}{arrangement} & \multicolumn{4}{c|}{others} & \multicolumn{4}{c}{\textbf{Total}} \\ \hline
\multirow{2}{*}{\textbf{Number of Categories}} & \multicolumn{1}{c|}{\textbf{r}} & \multicolumn{1}{c|}{\textbf{c}} & \multicolumn{1}{c|}{\textbf{f}} & \textbf{total} & \multicolumn{1}{c|}{\textbf{r}} & \multicolumn{1}{c|}{\textbf{c}} & \multicolumn{1}{c|}{\textbf{f}} & \textbf{total} & \multicolumn{1}{c|}{\textbf{r}} & \multicolumn{1}{c|}{\textbf{c}} & \multicolumn{1}{c|}{\textbf{f}} & \textbf{total} \\ \cline{2-13} 
 & \multicolumn{1}{c|}{0} & \multicolumn{1}{c|}{2} & \multicolumn{1}{c|}{1} & 3 & \multicolumn{1}{c|}{15} & \multicolumn{1}{c|}{20} & \multicolumn{1}{c|}{30} & 65 & \multicolumn{1}{c|}{337} & \multicolumn{1}{c|}{461} & \multicolumn{1}{c|}{405} & 1203 \\ \hline
\end{tabular}
\end{adjustbox}
\end{minipage}
\caption{Super classes and the number of categories in each super class. We list the number of \textbf{r}are, \textbf{c}ommon, and \textbf{f}requent categories under each super class.}
\label{tab:sc}
\end{table}

\subsection{Noise Label Generation}
\label{sec:nlg}

To add asymmetry (class-related) noise into the training set of the LVIS v1, we randomly pick a new category under the same super class to replace the original label. But, to make the semantics consistency between the original category and the new one, we only consider categories that do not belong to the ``others''. In Algorithm~\ref{alg:nlg}, we describe the process of generating label noise. The Clean Annotation List $A$ is a list containing all instances' categories, which has the following format: $A=[3,5,100,102,\dots,90,1000]$. $A[i]$ represent the $i$-th element in the list $A$. The Super Class Dictionary $S$ is a dictionary, whose key is the name of the super class name listed in Table~\ref{tab:sc}, and the value is a list of the categories which belong to the super class, e.g., $S=\{`\mathrm{instrumentality}':[1,3,10,\dots,900,1200],`\mathrm{food}':[2,7,\dots,1000, 1001],\dots,`\mathrm{others}':[90, 500, \dots,1203]\}$. We randomly sample a value from a uniform distribution $\mathcal{U}(0,1)$ between 0 and 1 to fit the Noise Ratio $p$. 

\begin{algorithm}[]
   \caption{Noise Label Generation}
   \label{alg:tail}
\begin{algorithmic}[1]
   \STATE {\bfseries Input:} Clean Annotation List $A$, Noise Ratio $p$, Super Class Dictionary $S$
   \FOR{$i = 0 \to (\mathrm{len}(A)-1)$}
   \STATE $r\gets \mathcal{U}(0,1)$
   \IF{$r \le p$}
   \STATE $s \gets $ super class of category $A[i]$
   \IF{$s$ is not ``others''}
   \STATE $n \gets $ randomly pick a category from $S[s]$
   \STATE $A[i] \gets n$
   \ENDIF
   \ENDIF
   \ENDFOR
   \STATE \textbf{return} $A$
\end{algorithmic}
\label{alg:nlg}
\end{algorithm}

\begin{table}[]
\centering
\begin{adjustbox}{max width=1.0\linewidth}
\begin{tabular}{c|c|c|c|c}
\hline
\textbf{\textbf{Method}} & \textbf{\textbf{Backbone}} & \textbf{\textbf{Normalized Mask}} & \textbf{\textbf{Normalized Classifier}} & \textbf{\textbf{Activation}} \\ \hline
EQL & \multirow{6}{*}{ResNet-50} & \multirow{6}{*}{\Checkmark} & \multirow{5}{*}{\XSolidBrush} & Sigmoid \\ \cline{1-1} \cline{5-5} 
EQL + Gumbel &  &  &  & Gumbel \\ \cline{1-1} \cline{5-5} 
DropLoss &  &  &  & Sigmoid \\ \cline{1-1} \cline{5-5} 
DropLoss + Gumbel &  &  &  & Gumbel \\ \cline{1-1} \cline{5-5} 
EQLv2 &  &  &  & Sigmoid \\ \cline{1-1} \cline{4-5} 
Seesaw Loss &  &  & \Checkmark & Softmax \\ \hline
\end{tabular}
\end{adjustbox}
\caption{Settings for different methods.}
\vspace{-10pt}
\label{tab:settings}
\end{table}

\begin{figure}[t] 
\centering
    \begin{subfigure}[b]{0.3\linewidth}
    \centering
    \includegraphics[width=\linewidth]{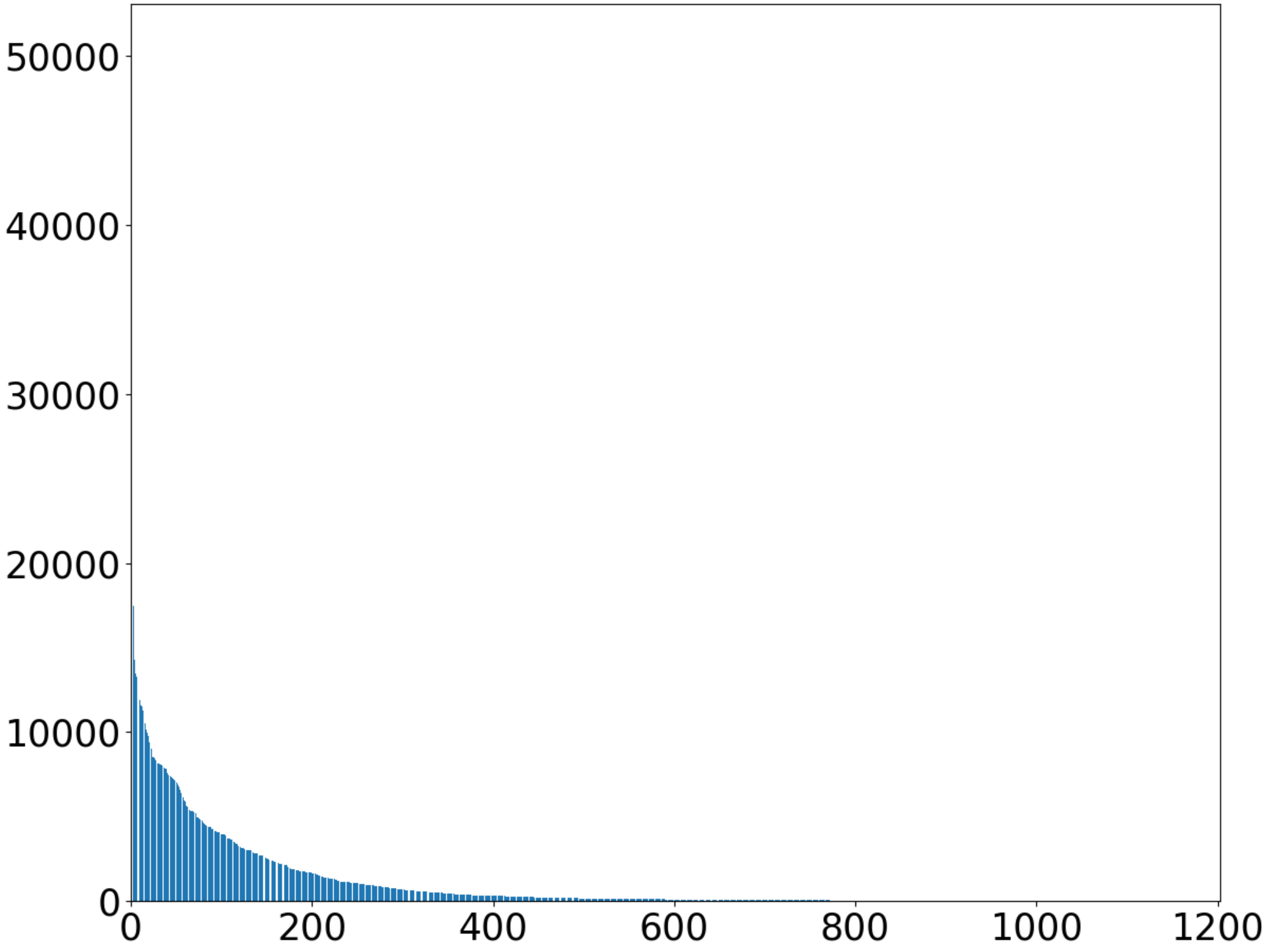} 
    \vspace{-15pt}
    \caption{LVIS v1 label distribution.}
    \label{fig:clean} 
  \end{subfigure} 
      \begin{subfigure}[b]{0.3\linewidth}
    \centering
    \includegraphics[width=\linewidth]{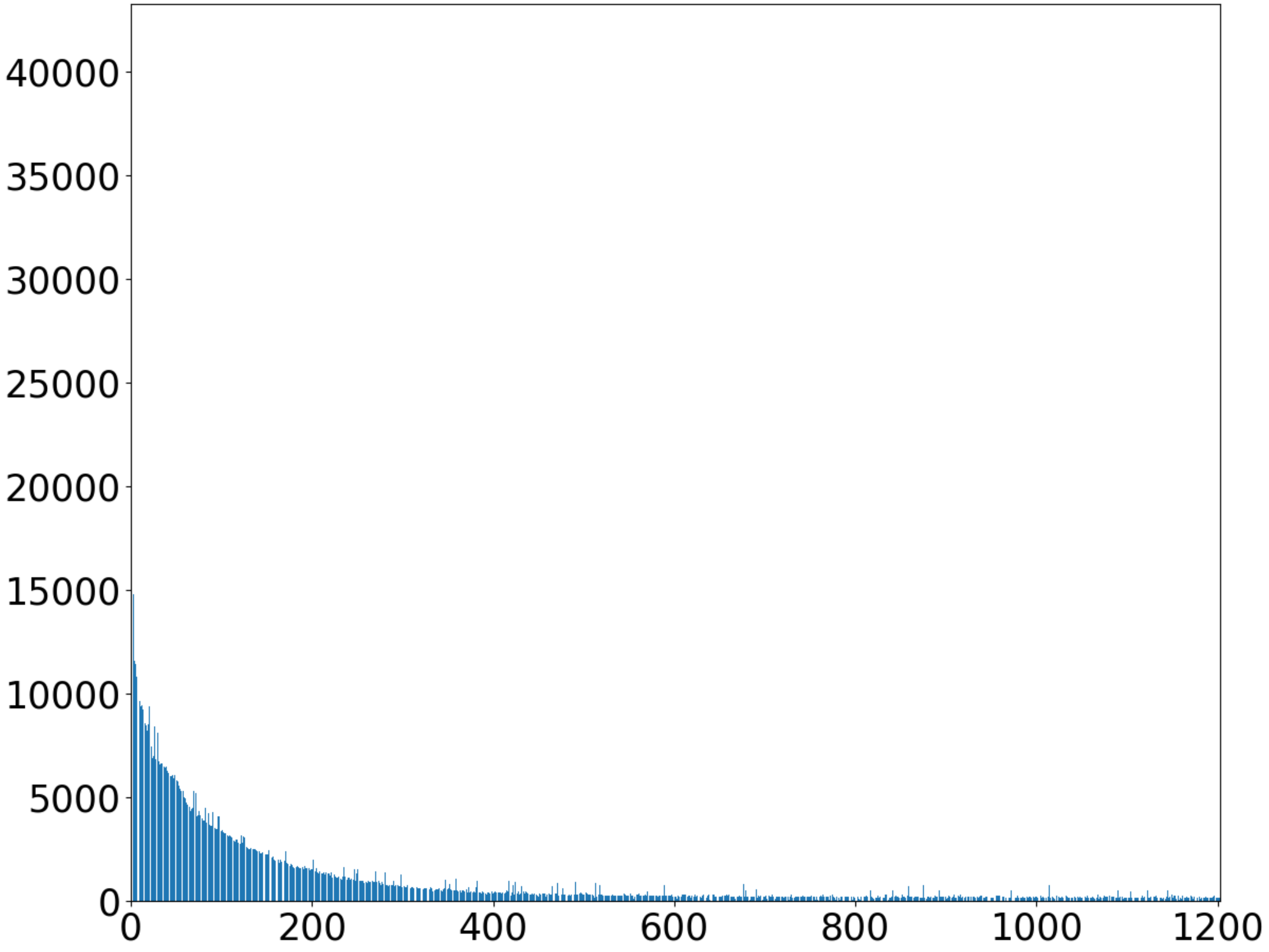}
    \vspace{-15pt}
    \caption{Asymmetric Noise (0.2).}
    \label{fig:an2} 
  \end{subfigure} 
        \begin{subfigure}[b]{0.3\linewidth}
    \centering
    \includegraphics[width=\linewidth]{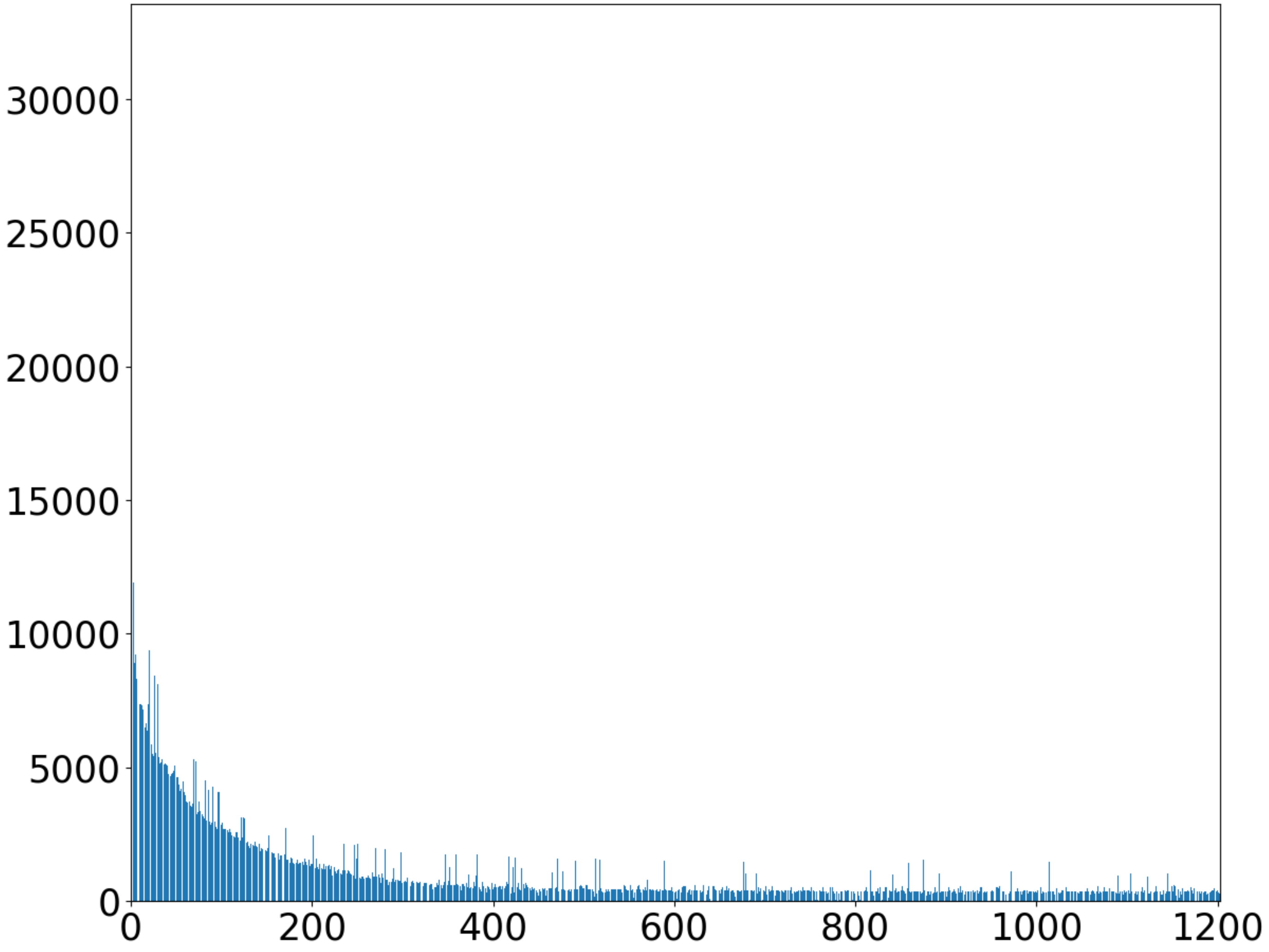}
    \vspace{-15pt}
    \caption{Asymmetric Noise (0.4).}
    \label{fig:an4} 
  \end{subfigure} \\
          \begin{subfigure}[b]{0.3\linewidth}
    \centering
    \includegraphics[width=\linewidth]{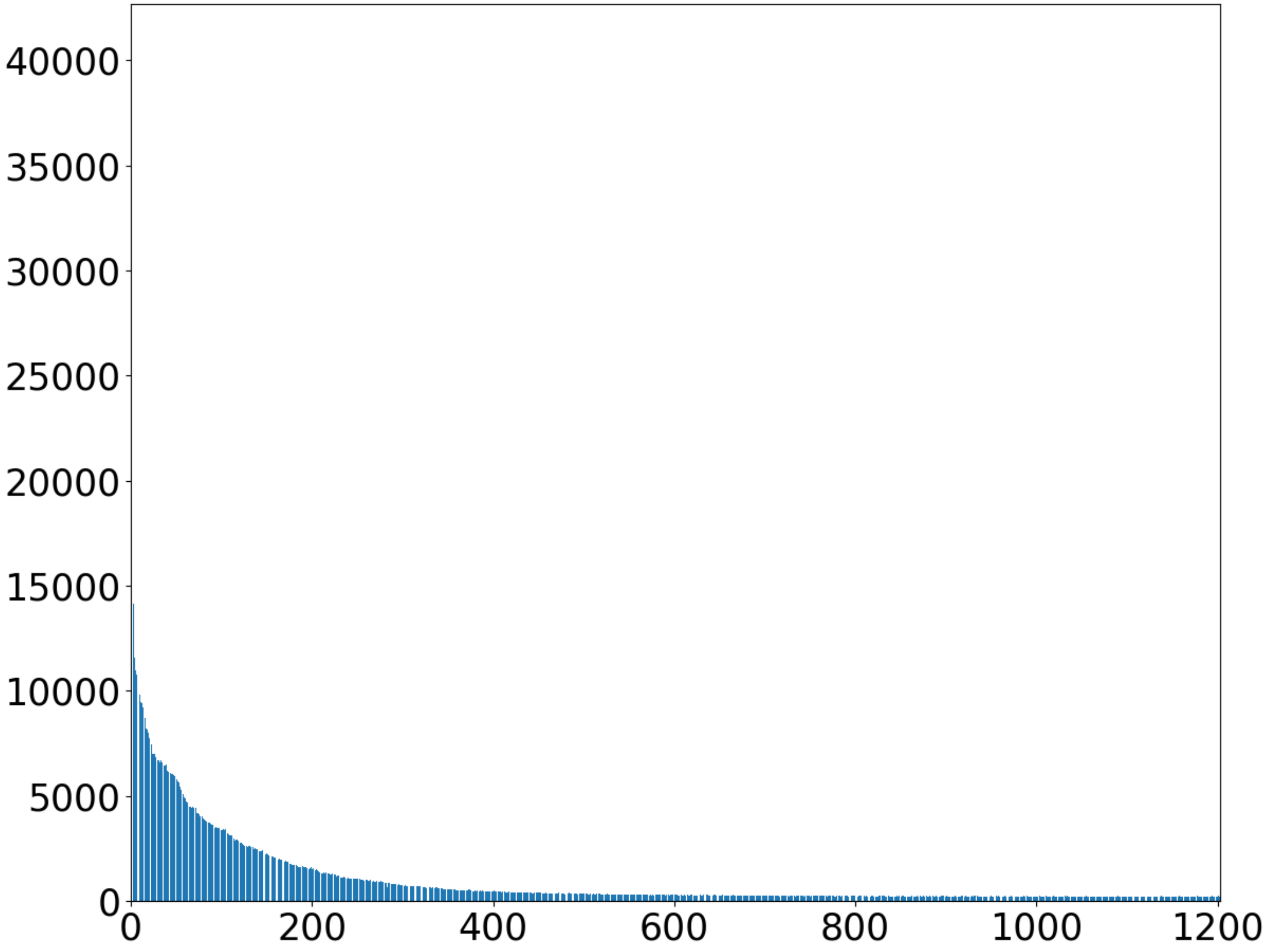}
    \vspace{-15pt}
    \caption{Symmetric Noise (0.2).}
    \label{fig:sn2} 
  \end{subfigure}
            \begin{subfigure}[b]{0.3\linewidth}
    \centering
    \includegraphics[width=\linewidth]{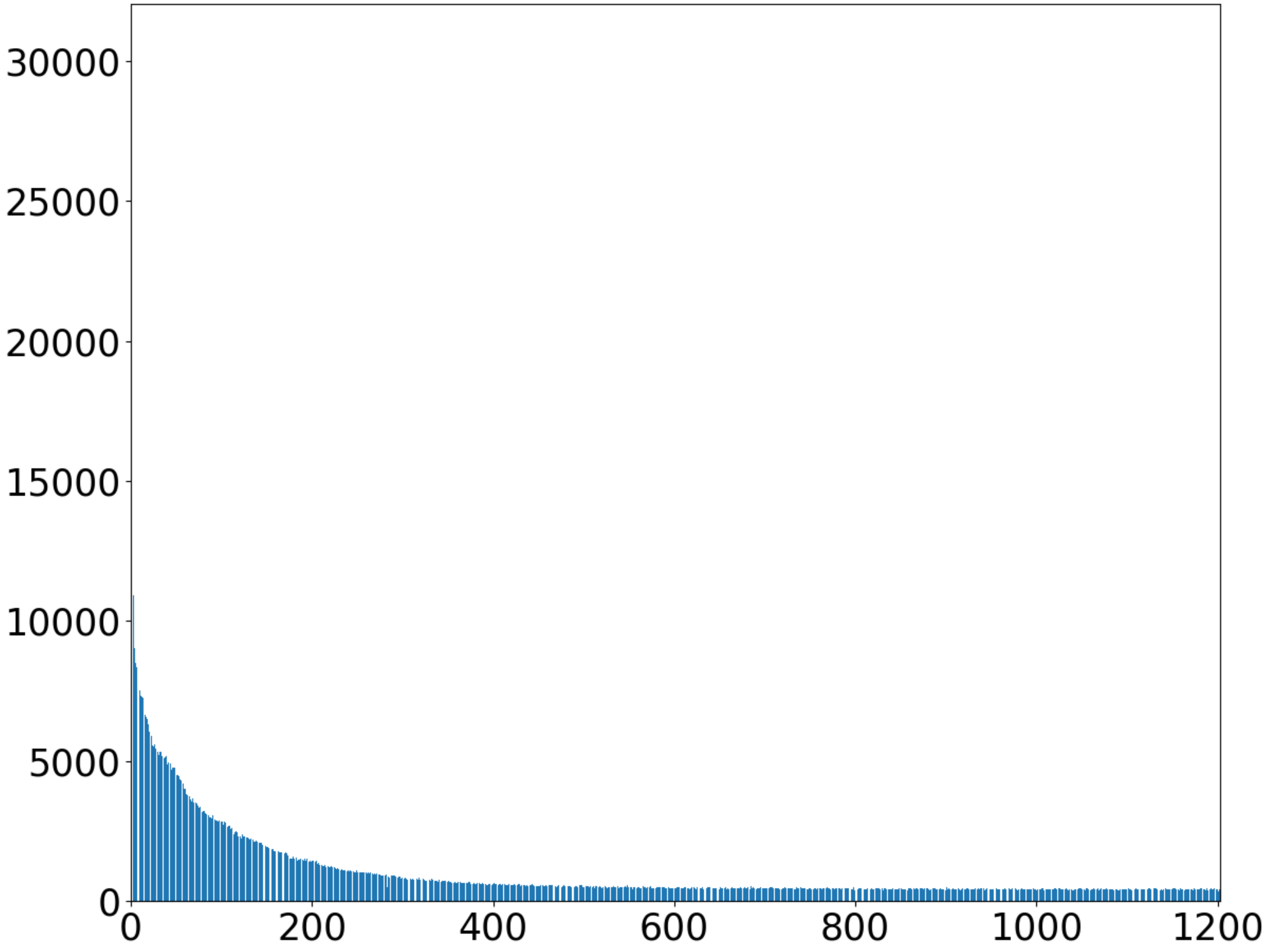}
    \vspace{-15pt}
    \caption{Symmetric Noise (0.4).}
    \label{fig:sn4} 
  \end{subfigure}
  \caption{Label distribution under different noise settings.}
  \label{fig:ansn} 
  \vspace{-20pt}
\end{figure}

\section{Experiments}
\label{sec:exp}

\subsection{Dataset}

In this paper, we adopt LVIS v1 to evaluate various methods' performance under different noise settings, which contains 100k images for training and 19.8k images for validation. In LVIS v1, there are 1,203 categories, grouped by their image frequency: rare (1-10 images in the dataset), common (11-100 images in the dataset) and frequent ($>$100 images in the dataset). For the label noise, we consider two types of the noise, i.e.,  symmetric (class-agnostic) noise and asymmetric (class-related) noise. When adding symmetric noise into the training set, we randomly pick a label from all 1,203 possible categories to replace the original label. When adding asymmetric noise, we follow the steps in Section~\ref{sec:nlg}. The results of the validation set are reported from five aspects: average segmentation performance $AP$, average box performance $AP^b$ and average segmentation performance for rare $AP^r$, common $AP^c$ and frequent categories $AP^f$.

\subsection{Methods}
We evaluate three mainstream long-tailed instance segmentation approaches, including EQL~\cite{tan_equalization_2020}, EQLv2~\cite{eqlv2}, DropLoss~\cite{hsieh_droploss_2021} and Seesaw Loss~\cite{wang_Seesaw_2021}. Specifically, the implementations of EQL and DropLoss are followed the Gumbel Optimized Loss~\cite{alexandridis_long-tailed_2022}, i.e., we replace the original activation function Sigmoid with Gumbel. 

\subsection{Implementation Details}

We adopt mmdetection~\cite{mmdetection} and Pytorch~\cite{pytorch} on 4 V100 GPU cards to implement all experiments. On each GPU card, there are 4 images in one batch. For all methods, we consider two model architectures, i.e., Mask R-CNN~\cite{he_mask_2017} and Cascade R-CNN~\cite{crcnn}, in which the ResNet-50~\cite{he_deep_2016} with FPN~\cite{fpn} is the backbone model. Furthermore, the $2\times$ schedule and the Normalized Mask~\cite{wang_Seesaw_2021} are adopted to improve the model's performance. The model's details are shown in Table~\ref{tab:settings}. For the training process, we adopt Stochastic Gradient Descent as our optimizer to train models with 24 epochs. The learning rate is 0.02, and we decrease it by 0.1 at the 16th epoch and the 22nd epoch. The data augmentations include random horizontal flipping and multi-scaling, which are consistent with previous works~\cite{wang_Seesaw_2021,alexandridis_long-tailed_2022}. In inference, we follow the protocol in~\cite{wang_Seesaw_2021}, i.e., we resize all images into 1333$\times$800 and use the score thresholds of $10^{-3}$. For the sampler, we consider the RFS sampler~\cite{gupta_lvis_2019} and the Random sampler, respectively.

\subsection{Results}

In Figure~\ref{fig:ansn}, we compare the distribution of the label in different noise settings. The $x$-axis is the category, which is sorted based on the frequency in the original training set of the LVIS v1. We keep the same category order in each image. And the $y$-axis is the number of instances that are annotated as the corresponding category. Clearly, adding noise will change the label distribution, thus the RFS sampler, which will over-sample low-frequency instances, will be influenced by the shifted label distribution. 

We first compare the results of Seesaw Loss with the RFS sampler and the Random sampler on Mask R-CNN in Table~\ref{tab:sampler}. For the RFS sampler, we set the over-sample threshold as 0.001. The results indicate that the RFS sampler can significantly increase the performance on the clean long-tailed dataset compared with the Random sampler. When there exist label noise in the dataset, the improvement obtained from the RFS sampler weakens, which can be the distribution shift, causing the sampler samples more instance with incorrect labels.

In Tables~\ref{tab:result} and \ref{tab:resultcrcnn}, we present the results under different loss functions on different models with different samplers. The results indicate that methods (DropLoss with Gumbel and EQL with Gumbel) with better performance may not be the best choice for practical when there exists noise in the training labels. On the other hand, different loss functions have different robustness for asymmetric (class-related) noise and symmetric (class-agnostic) noise. For example, the Seesaw Loss shows more robustness under symmetric noise. However, the DropLoss with Gumbel is more robust under asymmetric noise. This phenomenon indicates that designing a loss for a long-tailed instance segmentation dataset with noise is challenging, and it is complex to outperform under all types of noise.

\section{Conclusion}

In this report, we introduce a tool to automatically analyze the categories in LVIS v1. With it, we add asymmetric noise to the clean training set to create noisy datasets with different noise ratios. Then, we evaluate three mainstream long-tailed instance segmentation loss functions on the noisy datasets containing asymmetric noise and symmetric noise, respectively. The results indicate that previous long-tailed loss functions lack robustness when facing noisy datasets, and it is important to study how to design new ones to better fit the label noise.

\begin{table}[]
\centering
\begin{adjustbox}{max width=1.0\linewidth}
\begin{tabular}{c|c|c|c|c|c|c}
\hline
\textbf{Noise} & \textbf{Sampler} & \textbf{$AP^b$} & \textbf{$AP^r$} & \textbf{$AP^c$} & \textbf{$AP^f$} & \textbf{$AP$} \\ \hline
\multirow{2}{*}{None} & Random & 25.3 & 18.0 & 25.3 & 30.9 & 26.3 \\ \cline{2-7} 
 & RFS & 27.2 & 17.8 & 27.1 & 31.4 & 27.2 \\ \hline
\multirow{2}{*}{AN (0.2)} & Random & 21.6 & 5.2 & 19.6 & 29.1 & 20.9 \\ \cline{2-7} 
 & RFS & 22.0 & 6.7 & 20.2 & 29.4 & 21.5 \\ \hline
\multirow{2}{*}{AN (0.4)} & Random & 18.3 & 3.9 & 14.5 & 27.1 & 17.6 \\ \cline{2-7} 
 & RFS & 18.1 & 2.3 & 14.7 & 27.4 & 17.6 \\ \hline
\multirow{2}{*}{SN (0.2)} & Random & 22.9 & 6.9 & 21.7 & 29.2 & 22.1 \\ \cline{2-7} 
 & RFS & 22.8 & 7.1 & 21.5 & 29.7 & 22.2 \\ \hline
\multirow{2}{*}{SN (0.4)} & Random & 19.2 & 3.6 & 16.1 & 27.7 & 18.5 \\ \cline{2-7} 
 & RFS & 19.2 & 2.7 & 16.4 & 27.9 & 18.6 \\ \hline
\end{tabular}
\end{adjustbox}
\caption{Results of Seesaw Loss with different samplers on Mask R-CNN.}
\label{tab:sampler}
\vspace{-20pt}
\end{table}

\begin{table}[]
\centering
\begin{adjustbox}{max width=1.0\linewidth}
\begin{tabular}{c|c|ccccc|ccccc}
\hline
\multirow{2}{*}{\textbf{Noise}} & \multirow{2}{*}{\textbf{Method}} & \multicolumn{5}{c|}{\textbf{Random Sampler}} & \multicolumn{5}{c}{\textbf{RFS}} \\ \cline{3-12} 
 &  & \multicolumn{1}{c|}{\textbf{$AP^b$}} & \multicolumn{1}{c|}{\textbf{$AP^r$}} & \multicolumn{1}{c|}{\textbf{$AP^c$}} & \multicolumn{1}{c|}{\textbf{$AP^f$}} & \textbf{$AP$} & \multicolumn{1}{c|}{\textbf{$AP^b$}} & \multicolumn{1}{c|}{\textbf{$AP^r$}} & \multicolumn{1}{c|}{\textbf{$AP^c$}} & \multicolumn{1}{c|}{\textbf{$AP^f$}} & \textbf{$AP$} \\ \hline
\multirow{6}{*}{None} & DropLoss & \multicolumn{1}{c|}{21.1} & \multicolumn{1}{c|}{3.9} & \multicolumn{1}{c|}{21.1} & \multicolumn{1}{c|}{27.7} & 20.7 & \multicolumn{1}{c|}{22.4} & \multicolumn{1}{c|}{14.1} & \multicolumn{1}{c|}{22.6} & \multicolumn{1}{c|}{26.3} & 22.6 \\ \cline{2-12} 
 & DropLoss + Gumbel & \multicolumn{1}{c|}{25.3} & \multicolumn{1}{c|}{15.3} & \multicolumn{1}{c|}{25.3} & \multicolumn{1}{c|}{29.6} & 25.2 & \multicolumn{1}{c|}{27.6} & \multicolumn{1}{c|}{19.9} & \multicolumn{1}{c|}{28.0} & \multicolumn{1}{c|}{30.4} & 27.5 \\ \cline{2-12} 
 & EQL & \multicolumn{1}{c|}{22.8} & \multicolumn{1}{c|}{4.0} & \multicolumn{1}{c|}{22.4} & \multicolumn{1}{c|}{30.5} & 22.4 & \multicolumn{1}{c|}{26.6} & \multicolumn{1}{c|}{18.5} & \multicolumn{1}{c|}{26.2} & \multicolumn{1}{c|}{30.7} & 26.6 \\ \cline{2-12} 
 & EQL + Gumbel & \multicolumn{1}{c|}{24.7} & \multicolumn{1}{c|}{12.6} & \multicolumn{1}{c|}{24.7} & \multicolumn{1}{c|}{31.1} & 25.1 & \multicolumn{1}{c|}{26.8} & \multicolumn{1}{c|}{17.3} & \multicolumn{1}{c|}{26.7} & \multicolumn{1}{c|}{31.2} & 26.9 \\ \cline{2-12} 
 & EQLv2 & \multicolumn{1}{c|}{26.4} & \multicolumn{1}{c|}{20.1} & \multicolumn{1}{c|}{25.4} & \multicolumn{1}{c|}{31.0} & 26.7 & \multicolumn{1}{c|}{26.2} & \multicolumn{1}{c|}{19.0} & \multicolumn{1}{c|}{25.5} & \multicolumn{1}{c|}{31.3} & 26.7 \\ \cline{2-12} 
 & SeeSaw Loss & \multicolumn{1}{c|}{25.3} & \multicolumn{1}{c|}{18.0} & \multicolumn{1}{c|}{25.3} & \multicolumn{1}{c|}{30.9} & 26.3 & \multicolumn{1}{c|}{27.2} & \multicolumn{1}{c|}{17.8} & \multicolumn{1}{c|}{27.1} & \multicolumn{1}{c|}{31.4} & 27.2 \\ \hline
\multirow{6}{*}{AN (0.2)} & DropLoss & \multicolumn{1}{c|}{15.1} & \multicolumn{1}{c|}{1.3} & \multicolumn{1}{c|}{10.9} & \multicolumn{1}{c|}{24.6} & 14.6 & \multicolumn{1}{c|}{14.4} & \multicolumn{1}{c|}{2.3} & \multicolumn{1}{c|}{10.6} & \multicolumn{1}{c|}{23.5} & 14.2 \\ \cline{2-12} 
 & DropLoss + Gumbel & \multicolumn{1}{c|}{18.9} & \multicolumn{1}{c|}{3.3} & \multicolumn{1}{c|}{16.0} & \multicolumn{1}{c|}{27.3} & 18.2 & \multicolumn{1}{c|}{18.9} & \multicolumn{1}{c|}{3.6} & \multicolumn{1}{c|}{16.0} & \multicolumn{1}{c|}{27.7} & 18.4 \\ \cline{2-12} 
 & EQL & \multicolumn{1}{c|}{18.4} & \multicolumn{1}{c|}{2.0} & \multicolumn{1}{c|}{14.4} & \multicolumn{1}{c|}{28.7} & 17.8 & \multicolumn{1}{c|}{18.8} & \multicolumn{1}{c|}{2.7} & \multicolumn{1}{c|}{15.2} & \multicolumn{1}{c|}{29.1} & 18.5 \\ \cline{2-12} 
 & EQL + Gumbel & \multicolumn{1}{c|}{19.6} & \multicolumn{1}{c|}{3.2} & \multicolumn{1}{c|}{15.9} & \multicolumn{1}{c|}{29.2} & 18.9 & \multicolumn{1}{c|}{20.0} & \multicolumn{1}{c|}{4.4} & \multicolumn{1}{c|}{16.3} & \multicolumn{1}{c|}{29.7} & 19.5 \\ \cline{2-12} 
 & EQLv2 & \multicolumn{1}{c|}{21.8} & \multicolumn{1}{c|}{5.7} & \multicolumn{1}{c|}{19.6} & \multicolumn{1}{c|}{29.9} & 21.2 & \multicolumn{1}{c|}{21.5} & \multicolumn{1}{c|}{5.5} & \multicolumn{1}{c|}{19.5} & \multicolumn{1}{c|}{30.2} & 21.3 \\ \cline{2-12} 
 & SeeSaw Loss & \multicolumn{1}{c|}{21.6} & \multicolumn{1}{c|}{5.2} & \multicolumn{1}{c|}{19.6} & \multicolumn{1}{c|}{29.1} & 20.9 & \multicolumn{1}{c|}{22.0} & \multicolumn{1}{c|}{6.7} & \multicolumn{1}{c|}{20.2} & \multicolumn{1}{c|}{29.4} & 21.5 \\ \hline
\multirow{6}{*}{AN (0.4)} & DropLoss & \multicolumn{1}{c|}{-} & \multicolumn{1}{c|}{-} & \multicolumn{1}{c|}{-} & \multicolumn{1}{c|}{-} & - & \multicolumn{1}{c|}{-} & \multicolumn{1}{c|}{-} & \multicolumn{1}{c|}{-} & \multicolumn{1}{c|}{-} & - \\ \cline{2-12} 
 & DropLoss + Gumbel & \multicolumn{1}{c|}{14.6} & \multicolumn{1}{c|}{2.0} & \multicolumn{1}{c|}{9.1} & \multicolumn{1}{c|}{24.8} & 14.0 & \multicolumn{1}{c|}{14.7} & \multicolumn{1}{c|}{2.3} & \multicolumn{1}{c|}{9.5} & \multicolumn{1}{c|}{25.1} & 14.3 \\ \cline{2-12} 
 & EQL & \multicolumn{1}{c|}{14.7} & \multicolumn{1}{c|}{1.3} & \multicolumn{1}{c|}{8.3} & \multicolumn{1}{c|}{26.2} & 14.1 & \multicolumn{1}{c|}{14.9} & \multicolumn{1}{c|}{2.0} & \multicolumn{1}{c|}{8.7} & \multicolumn{1}{c|}{26.5} & 14.5 \\ \cline{2-12} 
 & EQL + Gumbel & \multicolumn{1}{c|}{15.6} & \multicolumn{1}{c|}{2.0} & \multicolumn{1}{c|}{9.4} & \multicolumn{1}{c|}{27.2} & 15.1 & \multicolumn{1}{c|}{15.9} & \multicolumn{1}{c|}{2.3} & \multicolumn{1}{c|}{9.6} & \multicolumn{1}{c|}{27.6} & 15.4 \\ \cline{2-12} 
 & EQLv2 & \multicolumn{1}{c|}{18.7} & \multicolumn{1}{c|}{2.7} & \multicolumn{1}{c|}{14.8} & \multicolumn{1}{c|}{28.5} & 18.0 & \multicolumn{1}{c|}{18.9} & \multicolumn{1}{c|}{3.1} & \multicolumn{1}{c|}{15.1} & \multicolumn{1}{c|}{28.8} & 18.4 \\ \cline{2-12} 
 & SeeSaw Loss & \multicolumn{1}{c|}{18.3} & \multicolumn{1}{c|}{3.9} & \multicolumn{1}{c|}{14.5} & \multicolumn{1}{c|}{27.1} & 17.6 & \multicolumn{1}{c|}{18.1} & \multicolumn{1}{c|}{2.3} & \multicolumn{1}{c|}{14.7} & \multicolumn{1}{c|}{27.4} & 17.6 \\ \hline
\multirow{6}{*}{SN (0.2)} & DropLoss & \multicolumn{1}{c|}{15.5} & \multicolumn{1}{c|}{0.9} & \multicolumn{1}{c|}{11.5} & \multicolumn{1}{c|}{25.0} & 15.0 & \multicolumn{1}{c|}{15.6} & \multicolumn{1}{c|}{1.3} & \multicolumn{1}{c|}{11.4} & \multicolumn{1}{c|}{25.3} & 15.1 \\ \cline{2-12} 
 & DropLoss + Gumbel & \multicolumn{1}{c|}{18.7} & \multicolumn{1}{c|}{2.7} & \multicolumn{1}{c|}{15.7} & \multicolumn{1}{c|}{27.5} & 18.1 & \multicolumn{1}{c|}{18.6} & \multicolumn{1}{c|}{3.5} & \multicolumn{1}{c|}{15.3} & \multicolumn{1}{c|}{27.9} & 18.2 \\ \cline{2-12} 
 & EQL & \multicolumn{1}{c|}{18.9} & \multicolumn{1}{c|}{2.1} & \multicolumn{1}{c|}{15.1} & \multicolumn{1}{c|}{29.0} & 18.3 & \multicolumn{1}{c|}{18.6} & \multicolumn{1}{c|}{1.6} & \multicolumn{1}{c|}{14.9} & \multicolumn{1}{c|}{29.1} & 18.1 \\ \cline{2-12} 
 & EQL + Gumbel & \multicolumn{1}{c|}{19.3} & \multicolumn{1}{c|}{2.9} & \multicolumn{1}{c|}{15.5} & \multicolumn{1}{c|}{29.1} & 18.7 & \multicolumn{1}{c|}{19.2} & \multicolumn{1}{c|}{3.3} & \multicolumn{1}{c|}{15.1} & \multicolumn{1}{c|}{29.5} & 18.7 \\ \cline{2-12} 
 & EQLv2 & \multicolumn{1}{c|}{22.3} & \multicolumn{1}{c|}{6.2} & \multicolumn{1}{c|}{20.2} & \multicolumn{1}{c|}{30.1} & 21.7 & \multicolumn{1}{c|}{22.4} & \multicolumn{1}{c|}{5.9} & \multicolumn{1}{c|}{20.5} & \multicolumn{1}{c|}{30.3} & 21.8 \\ \cline{2-12} 
 & SeeSaw Loss & \multicolumn{1}{c|}{22.9} & \multicolumn{1}{c|}{6.9} & \multicolumn{1}{c|}{21.7} & \multicolumn{1}{c|}{29.2} & 22.1 & \multicolumn{1}{c|}{22.8} & \multicolumn{1}{c|}{7.1} & \multicolumn{1}{c|}{21.5} & \multicolumn{1}{c|}{29.7} & 22.2 \\ \hline
\multirow{6}{*}{SN (0.4)} & DropLoss & \multicolumn{1}{c|}{-} & \multicolumn{1}{c|}{-} & \multicolumn{1}{c|}{-} & \multicolumn{1}{c|}{-} & - & \multicolumn{1}{c|}{-} & \multicolumn{1}{c|}{-} & \multicolumn{1}{c|}{-} & \multicolumn{1}{c|}{-} & - \\ \cline{2-12} 
 & DropLoss + Gumbel & \multicolumn{1}{c|}{13.8} & \multicolumn{1}{c|}{1.2} & \multicolumn{1}{c|}{7.4} & \multicolumn{1}{c|}{24.7} & 13.1 & \multicolumn{1}{c|}{13.8} & \multicolumn{1}{c|}{1.4} & \multicolumn{1}{c|}{7.5} & \multicolumn{1}{c|}{25.0} & 13.3 \\ \cline{2-12} 
 & EQL & \multicolumn{1}{c|}{14.2} & \multicolumn{1}{c|}{1.0} & \multicolumn{1}{c|}{7.3} & \multicolumn{1}{c|}{26.2} & 13.6 & \multicolumn{1}{c|}{14.3} & \multicolumn{1}{c|}{0.9} & \multicolumn{1}{c|}{7.5} & \multicolumn{1}{c|}{26.6} & 13.8 \\ \cline{2-12} 
 & EQL + Gumbel & \multicolumn{1}{c|}{14.6} & \multicolumn{1}{c|}{1.2} & \multicolumn{1}{c|}{7.4} & \multicolumn{1}{c|}{26.9} & 14.0 & \multicolumn{1}{c|}{14.5} & \multicolumn{1}{c|}{1.1} & \multicolumn{1}{c|}{7.4} & \multicolumn{1}{c|}{27.0} & 14.0 \\ \cline{2-12} 
 & EQLv2 & \multicolumn{1}{c|}{19.0} & \multicolumn{1}{c|}{2.8} & \multicolumn{1}{c|}{15.5} & \multicolumn{1}{c|}{28.5} & 18.4 & \multicolumn{1}{c|}{19.2} & \multicolumn{1}{c|}{3.5} & \multicolumn{1}{c|}{15.4} & \multicolumn{1}{c|}{29.0} & 18.7 \\ \cline{2-12} 
 & SeeSaw Loss & \multicolumn{1}{c|}{19.2} & \multicolumn{1}{c|}{3.6} & \multicolumn{1}{c|}{16.1} & \multicolumn{1}{c|}{27.7} & 18.5 & \multicolumn{1}{c|}{19.2} & \multicolumn{1}{c|}{2.7} & \multicolumn{1}{c|}{16.4} & \multicolumn{1}{c|}{27.9} & 18.6 \\ \hline
\end{tabular}
\end{adjustbox}
\caption{Results under different noise settings on Mask R-CNN.}
\label{tab:result}
\vspace{-20pt}
\end{table}

\begin{table}[]
\centering
\begin{adjustbox}{max width=1.0\linewidth}
\begin{tabular}{c|c|ccccc|ccccc}
\hline
\multirow{2}{*}{\textbf{Noise}} & \multirow{2}{*}{\textbf{Method}} & \multicolumn{5}{c|}{\textbf{Random Sampler}} & \multicolumn{5}{c}{\textbf{RFS}} \\ \cline{3-12} 
 &  & \multicolumn{1}{c|}{\textbf{$AP^b$}} & \multicolumn{1}{c|}{\textbf{$AP^r$}} & \multicolumn{1}{c|}{\textbf{$AP^c$}} & \multicolumn{1}{c|}{\textbf{$AP^f$}} & \textbf{$AP$} & \multicolumn{1}{c|}{\textbf{$AP^b$}} & \multicolumn{1}{c|}{\textbf{$AP^r$}} & \multicolumn{1}{c|}{\textbf{$AP^c$}} & \multicolumn{1}{c|}{\textbf{$AP^f$}} & \textbf{$AP$} \\ \hline
\multirow{4}{*}{None} & EQL & \multicolumn{1}{c|}{25.8} & \multicolumn{1}{c|}{4.3} & \multicolumn{1}{c|}{23.7} & \multicolumn{1}{c|}{31.1} & 23.2 & \multicolumn{1}{c|}{29.4} & \multicolumn{1}{c|}{16.9} & \multicolumn{1}{c|}{26.8} & \multicolumn{1}{c|}{31.2} & 26.8 \\ \cline{2-12} 
 & EQL + Gumbel & \multicolumn{1}{c|}{30.2} & \multicolumn{1}{c|}{16.6} & \multicolumn{1}{c|}{27.0} & \multicolumn{1}{c|}{31.6} & 27.0 & \multicolumn{1}{c|}{30.8} & \multicolumn{1}{c|}{19.4} & \multicolumn{1}{c|}{28.2} & \multicolumn{1}{c|}{31.7} & 28.0 \\ \cline{2-12} 
 & EQLv2 & \multicolumn{1}{c|}{30.6} & \multicolumn{1}{c|}{19.7} & \multicolumn{1}{c|}{26.4} & \multicolumn{1}{c|}{31.8} & 27.4 & \multicolumn{1}{c|}{30.2} & \multicolumn{1}{c|}{19.3} & \multicolumn{1}{c|}{27.3} & \multicolumn{1}{c|}{31.9} & 27.7 \\ \cline{2-12} 
 & SeeSaw Loss & \multicolumn{1}{c|}{31.3} & \multicolumn{1}{c|}{20.5} & \multicolumn{1}{c|}{27.5} & \multicolumn{1}{c|}{32.0} & 28.1 & \multicolumn{1}{c|}{31.0} & \multicolumn{1}{c|}{18.4} & \multicolumn{1}{c|}{28.4} & \multicolumn{1}{c|}{32.4} & 28.2 \\ \hline
\multirow{4}{*}{AN (0.2)} & EQL & \multicolumn{1}{c|}{20.1} & \multicolumn{1}{c|}{2.0} & \multicolumn{1}{c|}{14.4} & \multicolumn{1}{c|}{29.3} & 18.1 & \multicolumn{1}{c|}{20.7} & \multicolumn{1}{c|}{2.6} & \multicolumn{1}{c|}{15.3} & \multicolumn{1}{c|}{29.6} & 18.7 \\ \cline{2-12} 
 & EQL + Gumbel & \multicolumn{1}{c|}{22.0} & \multicolumn{1}{c|}{3.5} & \multicolumn{1}{c|}{17.1} & \multicolumn{1}{c|}{29.9} & 19.7 & \multicolumn{1}{c|}{22.4} & \multicolumn{1}{c|}{4.1} & \multicolumn{1}{c|}{17.6} & \multicolumn{1}{c|}{30.3} & 20.3 \\ \cline{2-12} 
 & EQLv2 & \multicolumn{1}{c|}{24.7} & \multicolumn{1}{c|}{5.9} & \multicolumn{1}{c|}{20.8} & \multicolumn{1}{c|}{30.8} & 22.1 & \multicolumn{1}{c|}{24.7} & \multicolumn{1}{c|}{6.1} & \multicolumn{1}{c|}{21.4} & \multicolumn{1}{c|}{31.0} & 22.5 \\ \cline{2-12} 
 & SeeSaw Loss & \multicolumn{1}{c|}{25.1} & \multicolumn{1}{c|}{6.8} & \multicolumn{1}{c|}{21.5} & \multicolumn{1}{c|}{30.3} & 22.4 & \multicolumn{1}{c|}{24.9} & \multicolumn{1}{c|}{5.9} & \multicolumn{1}{c|}{21.7} & \multicolumn{1}{c|}{30.8} & 22.5 \\ \hline
\multirow{4}{*}{AN (0.4)} & EQL & \multicolumn{1}{c|}{15.7} & \multicolumn{1}{c|}{1.2} & \multicolumn{1}{c|}{7.7} & \multicolumn{1}{c|}{26.7} & 14.0 & \multicolumn{1}{c|}{16.0} & \multicolumn{1}{c|}{1.6} & \multicolumn{1}{c|}{8.3} & \multicolumn{1}{c|}{26.9} & 14.4 \\ \cline{2-12} 
 & EQL + Gumbel & \multicolumn{1}{c|}{17.4} & \multicolumn{1}{c|}{2.2} & \multicolumn{1}{c|}{9.8} & \multicolumn{1}{c|}{27.7} & 15.5 & \multicolumn{1}{c|}{17.5} & \multicolumn{1}{c|}{2.0} & \multicolumn{1}{c|}{10.3} & \multicolumn{1}{c|}{28.1} & 15.9 \\ \cline{2-12} 
 & EQLv2 & \multicolumn{1}{c|}{21.1} & \multicolumn{1}{c|}{2.7} & \multicolumn{1}{c|}{15.6} & \multicolumn{1}{c|}{29.4} & 18.8 & \multicolumn{1}{c|}{21.3} & \multicolumn{1}{c|}{2.8} & \multicolumn{1}{c|}{16.2} & \multicolumn{1}{c|}{29.7} & 19.2 \\ \cline{2-12} 
 & SeeSaw Loss & \multicolumn{1}{c|}{20.9} & \multicolumn{1}{c|}{4.2} & \multicolumn{1}{c|}{15.5} & \multicolumn{1}{c|}{28.6} & 18.7 & \multicolumn{1}{c|}{21.1} & \multicolumn{1}{c|}{3.7} & \multicolumn{1}{c|}{16.1} & \multicolumn{1}{c|}{29.0} & 19.0 \\ \hline
\multirow{4}{*}{SN (0.2)} & EQL & \multicolumn{1}{c|}{20.1} & \multicolumn{1}{c|}{1.3} & \multicolumn{1}{c|}{14.5} & \multicolumn{1}{c|}{29.4} & 18.0 & \multicolumn{1}{c|}{20.4} & \multicolumn{1}{c|}{1.5} & \multicolumn{1}{c|}{14.9} & \multicolumn{1}{c|}{29.8} & 18.4 \\ \cline{2-12} 
 & EQL + Gumbel & \multicolumn{1}{c|}{21.3} & \multicolumn{1}{c|}{3.4} & \multicolumn{1}{c|}{15.8} & \multicolumn{1}{c|}{29.7} & 19.1 & \multicolumn{1}{c|}{21.4} & \multicolumn{1}{c|}{3.5} & \multicolumn{1}{c|}{16.0} & \multicolumn{1}{c|}{30.1} & 19.4 \\ \cline{2-12} 
 & EQLv2 & \multicolumn{1}{c|}{25.6} & \multicolumn{1}{c|}{7.2} & \multicolumn{1}{c|}{22.2} & \multicolumn{1}{c|}{30.8} & 23.0 & \multicolumn{1}{c|}{25.2} & \multicolumn{1}{c|}{6.0} & \multicolumn{1}{c|}{22.2} & \multicolumn{1}{c|}{31.1} & 22.9 \\ \cline{2-12} 
 & SeeSaw Loss & \multicolumn{1}{c|}{26.1} & \multicolumn{1}{c|}{7.0} & \multicolumn{1}{c|}{23.1} & \multicolumn{1}{c|}{30.5} & 23.2 & \multicolumn{1}{c|}{25.9} & \multicolumn{1}{c|}{6.9} & \multicolumn{1}{c|}{23.1} & \multicolumn{1}{c|}{30.8} & 23.3 \\ \hline
\multirow{4}{*}{SN (0.4)} & EQL & \multicolumn{1}{c|}{14.9} & \multicolumn{1}{c|}{0.7} & \multicolumn{1}{c|}{6.4} & \multicolumn{1}{c|}{26.4} & 13.2 & \multicolumn{1}{c|}{15.0} & \multicolumn{1}{c|}{0.7} & \multicolumn{1}{c|}{6.5} & \multicolumn{1}{c|}{26.6} & 13.4 \\ \cline{2-12} 
 & EQL + Gumbel & \multicolumn{1}{c|}{15.8} & \multicolumn{1}{c|}{0.8} & \multicolumn{1}{c|}{7.5} & \multicolumn{1}{c|}{27.2} & 14.1 & \multicolumn{1}{c|}{15.9} & \multicolumn{1}{c|}{0.9} & \multicolumn{1}{c|}{7.7} & \multicolumn{1}{c|}{27.6} & 14.3 \\ \cline{2-12} 
 & EQLv2 & \multicolumn{1}{c|}{21.9} & \multicolumn{1}{c|}{3.4} & \multicolumn{1}{c|}{16.8} & \multicolumn{1}{c|}{29.7} & 19.5 & \multicolumn{1}{c|}{21.9} & \multicolumn{1}{c|}{2.8} & \multicolumn{1}{c|}{17.2} & \multicolumn{1}{c|}{29.9} & 19.7 \\ \cline{2-12} 
 & SeeSaw Loss & \multicolumn{1}{c|}{22.1} & \multicolumn{1}{c|}{3.6} & \multicolumn{1}{c|}{17.6} & \multicolumn{1}{c|}{29.1} & 19.7 & \multicolumn{1}{c|}{22.5} & \multicolumn{1}{c|}{4.2} & \multicolumn{1}{c|}{18.1} & \multicolumn{1}{c|}{29.7} & 20.2 \\ \hline
\end{tabular}
\end{adjustbox}
\caption{Results under different noise settings on Cascade R-CNN.}
\label{tab:resultcrcnn}
\vspace{-20pt}
\end{table}

\clearpage

\begin{center}
    \adjincludegraphics[height=500pt, trim={{.8\width} 0 {.1\width} 0}, clip, angle=90]{example1_graph.pdf}
    \adjincludegraphics[height=500pt, trim={{.7\width} 0 {.2\width} 0}, clip, angle=90]{example1_graph.pdf}
    \adjincludegraphics[height=500pt, trim={{.6\width} 0 {.3\width} 0}, clip, angle=90]{example1_graph.pdf}
    \adjincludegraphics[height=500pt, trim={{.5\width} 0 {.4\width} 0}, clip, angle=90]{example1_graph.pdf}
    \adjincludegraphics[height=500pt, trim={{.4\width} 0 {.5\width} 0}, clip, angle=90]{example1_graph.pdf}
    \adjincludegraphics[height=500pt, trim={{.3\width} 0 {.6\width} 0}, clip, angle=90]{example1_graph.pdf}
    \adjincludegraphics[height=500pt, trim={{.2\width} 0 {.7\width} 0}, clip, angle=90]{example1_graph.pdf}
    \adjincludegraphics[height=500pt, trim={{.1\width} 0 {.8\width} 0}, clip, angle=90]{example1_graph.pdf}
    \adjincludegraphics[height=500pt, trim={0 0 {.9\width} 0}, clip, angle=90]{example1_graph.pdf}
\end{center}
\bibliographystyle{alpha}
\bibliography{bib}

\end{document}